\pdfoutput=1

\documentclass[10pt, a4paper]{article}

\usepackage{lrec-coling2024} 

\usepackage{times}
\usepackage{latexsym}

\usepackage[T2A,T1]{fontenc}
\usepackage[russian,english]{babel}
\usepackage[utf8]{inputenc}

\usepackage{microtype}

\usepackage{tabularx}
\usepackage{booktabs}
\usepackage{adjustbox}
\usepackage[capitalise,noabbrev,nameinlink]{cleveref}

\title{Enriching Word Usage Graphs with Cluster Definitions}

\name{Mariia Fedorova\textsuperscript{1}, Andrey Kutuzov\textsuperscript{1}, Nikolay Arefyev\textsuperscript{1}, Dominik Schlechtweg\textsuperscript{2}} 

\address{\textsuperscript{1}University of Oslo, \textsuperscript{2}University of Stuttgart \\
         \{mariiaf, andreku, nikolare\}@ifi.uio.no\\
         dominik.schlechtweg@ims.uni-stuttgart.de\\}

\abstract{
We present a dataset of word usage graphs (WUGs), where the existing WUGs for multiple languages are enriched with cluster labels functioning as sense definitions. They are generated from scratch by fine-tuned encoder-decoder language models. The conducted human evaluation has shown that these definitions match the existing clusters in WUGs better than the definitions chosen from WordNet by two baseline systems. At the same time, the method is straightforward to use and easy to extend to new languages. The resulting enriched datasets can be extremely helpful for moving on to explainable semantic change modeling.
 \\ \newline  }

\begin{document}
\maketitleabstract

\section{Introduction and related work}

In natural languages most words are polysemous, i.e. the same word in different contexts can have different meanings. This gave rise to NLP tasks like word sense induction and word sense disambiguation. To solve them, the NLP community came up with many valuable resources related to word senses. One specific use case for such datasets is lexical semantic change modeling, where diachronic changes of word meaning are traced in an automated way. 

The field of semantic change modeling makes a heavy use of the so called `Word Usage Graphs' (WUGs) \cite{schlechtweg-etal-2021-dwug}. Each WUG is associated with a particular target word, and is a weighted, undirected graph, with nodes corresponding to target word usages in a corpus, while edges are weighted with the semantic proximity of particular pairs of usages.  WUGs are human-annotated, with annotators yielding contextualized graded judgments about the said semantic proximity. After the annotation is complete, graph nodes (target word occurrences) are automatically clustered into groups roughly corresponding to word senses.  When a WUG contains target word occurrences from different time periods, it is called a \textit{Diachronic} Word Usage Graph (DWUG).

DWUGs are available for several languages and are often used for evaluating semantic change detection and discovery systems \cite{schlechtweg-etal-2020-semeval, kutuzov-etal-2022-contextualized}.  However, one of the features they lack is human interpretability of senses: clusters are labeled only with their numerical identifiers. To analyze a cluster from a DWUG, one has to browse through the actual target word usages, which is cumbersome and time-consuming. \citet{giulianelli-etal-2023-interpretable} addressed this issue and suggested a method to automatically generate human-readable sense definitions (cluster labels) using an encoder-decoder language model fine-tuned on the task of contextualized definition generation \cite{mickus-etal-2022-semeval}. However, their work was limited to the English DWUGs and they did not release the labels they produced.

In this paper, we apply this method to actually generate sense definitions for the available diachronic and synchronic WUGs in several languages (English, German, Norwegian, and Russian). We publicly release these enriched WUGs (mappings from clusters to definitions) with definitions both in English and in the WUG-specific languages.\footnote{\url{https://github.com/ltgoslo/wugs_with_definitions}} In addition, we compare the performance of the LLM-based definition generation system with other approaches which choose a definition from an existing ontology like Wordnet or Wiktionary. Thus, our contribution is twofold: first, we make the existing word usage graphs more useful for linguists and lexicographers; second, we evaluate definition generation and aggregation methods on multiple languages and release the best fine-tuned models.

\section{Data Description}

\subsection{Word usage graphs}

The WUG repository \citelanguageresource{wugs} features word usage graphs for eight languages. We chose English, German, Norwegian and Russian to experiment with. We ignore Chinese, Latin, Spanish and Swedish for the time being: mostly because of the lack of available datasets to fine-tune definition generators for these languages.

\begin{table*}
\centering
\begin{tabular}{@{}l|llll@{}}
\toprule
\textbf{Language} & \textbf{Target words} & \textbf{Clusters} & \textbf{Clusters annotated} & \textbf{Diachronic?}\\
\midrule
English \cite{schlechtweg-etal-2020-semeval} & 46 & 819 & 120 & True \\
German \cite{schlechtweg-etal-2020-semeval} & 50 & 488 & 95 & True \\
Norwegian-1 \cite{kutuzov-etal-2022-nordiachange} & 40 & 99 & 23 & True \\
Norwegian-2 \cite{kutuzov-etal-2022-nordiachange} & 40 & 78 & 17 & True \\
Russian \cite{aksenova-etal-2022-rudsi} & 24 & 90 & 39 & False \\
\bottomrule
\end{tabular}
\caption{Main statistics of the enriched WUGs. We labeled and annotated only clusters which feature at least three usage examples.}
\label{tab:dwugs}
\end{table*}

Table~\ref{tab:dwugs} provides the main statistics of the word usage graphs we employ. Note that there are \textit{two} Norwegian DWUGs with different target word sets and time periods (Norwegian-1 and Norwegian-2). Also note that for Russian we use the `\textit{RuDSI}' WUGs \cite{aksenova-etal-2022-rudsi} which are not diachronic (target word usages may come from different time period, but it is not specially accounted for). There exist two semantic change detection datasets for Russian (`\textit{RuSemShift}' and `\textit{RuShiftEval}'), but due to their annotation procedure, they do not feature any meaningful clusters to label.

We did not produce labels for clusters with less than three example usages (this excluded a large amount of singleton clusters). The reason is that many of such `small' groups are clustering errors, and even if they are not, it is extremely difficult to come with a good definition based on one or two examples; for the DefGen method (subsection~\ref{subsec:defgen}), generating a definition for a cluster with two examples is ill-defined. We leave labeling this long tail of small clusters to future work.\footnote{Also, in some original DWUGs, there are clusters labeled with `-1', which means annotators were unsure about semantic proximity for nodes within these clusters most of the time; we ignored these clusters as well.}

\subsection{Definition datasets}
\label{subsec:defin_datasets}

To be able to generate contextualized dictionary-like definitions, a pre-trained encoder-decoder language model has to be fine-tuned on a dataset of definitions coupled to example target word usages. In addition, one has to choose a prompt which serves as an instruction for the language model to do the correct task. We follow the logic of \citet{giulianelli-etal-2023-interpretable} in choosing both the prompt (`\textit{What is the definition of TARGET\_WORD?}' and its translations into corresponding languages) and the fine-tuning datasets for English: \textbf{WordNet}~\cite{ishiwatari-etal-2019-learning}, \textbf{Oxford}~\cite{gadetsky-etal-2018-conditional} and \textbf{CoDWoE}~\cite{mickus-etal-2022-semeval} (all CoDWoE datasets originally come from Wiktionary). For other languages we used the following resources:
\begin{itemize}
    \item Russian: the corresponding part of \textbf{CoDWoE};
    \item Norwegian: \textbf{Bokmålsordboka}~\citelanguageresource{Ordbokene}.
\end{itemize}

We are not aware of any readily available definition dataset for German; however, we still decided to assign cluster labels to German DWUGs as an experiment in zero-shot cross-lingual definition generation (also, we had available resources for human evaluation of German cluster labels). See \ref{sec:generation} for more details on definition generation for different languages. 

Table~\ref{tab:definition_datasets} shows the statistics of the definition datasets for fine-tuning.

\begin{table*}
\centering

\begin{tabular}{@{}l|lccll@{}}
\toprule
\textbf{Dataset} & \textbf{Entries} & \textbf{Lemmas} & \textbf{Ratio} & \textbf{Usage Length} & \textbf{Definition Length} \\
\midrule
\textbf{English} & 175 332 & 49 238 & 3.56 & $31.64^{\pm23.82}$ & $17.56^{\pm11.13}$ \\
\textbf{Norwegian} & 70 711 & 30 551 & 2.31 & $8.69^{\pm3.40}$ & $12.25^{\pm7.35}$ \\
\textbf{Russian} & 72 872 & 34 495 & 2.11 & $52.03^{\pm29.37}$ & $24.65^{\pm15.84}$ \\
\bottomrule
\end{tabular}

\caption{Definition datasets. Average usage and definition lengths are given in \texttt{mT0} sub-words.}
\label{tab:definition_datasets}
\end{table*}

\section{Definition generators}
\label{sec:generation}
We present three methods that can enrich DWUG clusters with definitions. Two of them are our baselines: they select a definition for each cluster from a human-curated lexical ontology, the English WordNet in our case. The third one generates definitions from scratch and is our main method.

\subsection{Lesk}
The simplest method uses the Lesk algorithm~\cite{lesk1986automatic} originally developed to solve the Word Sense Disambiguation (WSD) task. For a target word in a given context, it browses all the definitions listed for this word in a glossary or some WordNet-like resource and picks up a definition with the highest lexical overlap with the given context. To construct a context for our task we concatenate all usages from the same cluster, thus selecting the most suitable definition for this particular cluster.
The NLTK \cite{bird2009natural} implementation
of the Lesk algorithm was employed, together with WordNet as the source of senses and their definitions. 
We used the pre-tokenized versions of WUG examples, while WordNet definitions were split into tokens by whitespaces. The part-of-speech tags of the target words (where available) were taken into account when selecting synsets.
This method is doomed to fail when the target word occurs in a sense that is not listed among senses of this target word in WordNet, or if the target word is simply absent in WordNet. Also, it does not work with languages other than English unless using lexical resources for those languages. In our experiments with the English DWUGs, we observed very low accuracy compared to other methods even for English which has well-developed lexical resources (see Section~\ref{sec:evaluation}). Thus, we decided not to adapt Lesk to other languages.

\subsection{GlossReader}
GlossReader~\citep{rachinskiy-arefyev-2021-mclwic} is a system originally developed to produce the contextualized embeddings that were shown to outperform standard LLM embeddings in lexical semantic tasks such as Multilingual and Cross-lingual Word-in-Context~\citep{martelli-etal-2021-semeval} and Lexical Semantic Change Discovery~\citep{d-zamora-reina-etal-2022-black}. The system is based on the English WSD model by \cite{blevins2020wsd}, which consists of a definition (gloss) encoder and a context encoder, both initialized with the English BERT weights and jointly fine-tuned on a WSD dataset to maximize the dot product between a context embedding and a definition embedding of the corresponding word sense. In GlossReader, the English BERT backbone was replaced by the multilingual \texttt{XLM-R} \cite{conneau-etal-2020-unsupervised} in both encoders, and it was shown that fine-tuning \texttt{XLM-R} as part of such WSD system significantly improves the performance of its contextualized embeddings for several lexical semantic tasks in various languages even when fine-tuning on the same English WSD dataset only~\citep{rachinskiy-arefyev-2021-mclwic,rachinskiy-arefyev-2022-lscdiscovery}.
In this work, we first build gloss embeddings for all WordNet definitions (about 117K) with the gloss encoder. Then, for each word usage we build the contextualized embedding of the target word and retrieve $k$ most similar definition embeddings\footnote{The pairwise human annotations from DWUGs were employed to select the optimal value of $k$ among 1,3,10. For each DWUG and each $k$ we built a definition for each cluster, then assigned cluster definitions to all of its usages and estimated the probability that two usages with entirely different senses (the annotation of 1) obtain the same definition. We selected $k=3$ for English DWUG and $k=10$ for other DWUGs to minimize this probability.} as measured by the dot product. Finally, for each cluster we select the definition retrieved for the largest number of usages in this cluster.

Unlike Lesk, this method selects from \textit{all} definitions in WordNet, thus, it can produce reasonable definitions even for senses not listed for the target word in WordNet. Moreover, due to zero-shot cross-lingual transferability of the \texttt{XLM-R} backbone, GlossReader can produce reasonable embeddings for definitions and word usages in various languages, and retrieve definitions in one language for word usages in another language. In this work, we use GlossReader to retrieve definitions in English for word usages in several other languages (cross-lingual setup), but we leave experiments with retrieving definitions in other languages (multilingual setup) for the future work.

\subsection{DefGen}
\label{subsec:defgen}
We dub `DefGen' our main system, which takes as an input a text string containing a target word usage, and generates a human-readable definition of the target word in this particular context from scratch. We use the method proposed by \citet{giulianelli-etal-2023-interpretable}: an encoder-decoder language model is fine-tuned on a dataset of target word usages and the corresponding definitions (see the subsection~\ref{subsec:defin_datasets}). Then, definitions are conditionally generated for every example in the test set (in our case, WUGs). \citet{giulianelli-etal-2023-interpretable} used \texttt{Flan-T5} \cite{chung2022scaling} as the underlying language model. However, it was trained predominantly on English and lacks the capability to properly encode or generate texts in languages with significantly different writing systems (especially true for Russian, but some German and Norwegian characters are also not processed by the \texttt{Flan-T5}  tokenizer).  Because of that, in this work we are using \texttt{mT0}~\cite{muennighoff-etal-2023-crosslingual} which is essentially a multilingual version of \texttt{Flan-T5}  (also fine-tuned on many datasets cast as natural language instructions). For all the experiments, we employ the \texttt{mT0-xl} version of the model\footnote{\url{https://huggingface.co/bigscience/mT0-xl}}, 3.7B parameters in size.

Fine-tuning was done in a standard text-to-text setup, for every language (English, Norwegian, Russian) separately, so that in the end we had three language-specific models. However, the underlying model is multilingual, so e.g., the English DefGen can still take sentences in other languages as input, and produce target word definitions in English with the standard English prompt. We used it to generate English definitions for the German DWUGs. Here, the input to the model looked like \textit{`Ist eine Prüfung erforderlich, so erfolgt eine Entscheidung über den Antrag durch die zuständige Behörde. What is the definition of Entscheidung?' (`If an examination is
necessary, a decision on the application will be made by the responsible authority. What is the definition of decision?')}\footnote{The \texttt{mT0} answer in this case was \textit{`The act of making up your mind about something; a decision.'}}. Note that the models were not specifically fine-tuned on producing English definitions for German examples: this is a zero-shot multilingual ability probably stemming from the large-scale instruction fine-tuning of the base model on other tasks. 

Table~\ref{tab:rougel} shows the raw performance of the fine-tuned models with English and language-specific (`native') prompts as ROUGE-L \cite{lin-2004-rouge} on the validation set. Note that here we report only the performance for definitions generated in the same language as the original examples (we don't have any datasets with examples in one language and definitions in another to evaluate against). ROUGE-L measures only surface form overlap, so it should not be considered as the definitive metric. Still, the performance of our definition generators for English and Norwegian is on par with the English results reported in \citet{giulianelli-etal-2023-interpretable}. The lower score on the Russian CodWoE dataset is most probably related to the abundance of dictionary-specific abbreviations, to morphological richness of Russian and to the fact that the reference Russian definitions are longer on average than those for other languages, as well as usage examples (Table~\ref{tab:definition_datasets}). This is why the lengths of the longest common sub-sequence present in the reference definition and the generated definition are lower on average. Still, the definitions generated by the Russian model are mostly relevant and meaningful, as confirmed with visual examination by native Russian speakers and with the human evaluation experiments in Section~\ref{sec:evaluation}. It is also worth noting that the performance when using English or language-specific prompts is not significantly different. In all the subsequent experiments, we use native prompts when generating language-specific definitions.

\begin{table}
\centering
\begin{tabular}{@{}l|ll@{}}
\toprule
\textbf{Language} & \textbf{English prompt} & \textbf{Native prompt} \\
\midrule
English & 39.14 & 39.14 \\
Norwegian & 28.16 & 27.76 \\
Russian & 17.26 & 17.25 \\
\bottomrule
\end{tabular}

\caption{Performance of \texttt{mT0}-based definition generators (ROUGE-L * 100) on the validation sets.}
\label{tab:rougel}
\end{table}

Once the definitions for all the examples of a specific WUG cluster in the desired language are generated, we again follow \citet{giulianelli-etal-2023-interpretable} in choosing \textit{the most prototypical definition} as the cluster label. All the generated definitions are encoded by a multilingual SBERT model\footnote{\url{https://huggingface.co/sentence-transformers/distiluse-base-multilingual-cased-v1}} \cite{reimers-gurevych-2019-sentence} and the most prototypical definition is trivially the one with the vector representation closest (by cosine similarity) to the average of all the definition embeddings in the cluster. No search for hyperparameters was conducted, we used the default values from \citet{giulianelli-etal-2023-interpretable} at all the DefGen stages.

In this way, we generated English cluster/sense labels for all four languages. In addition, language-specific labels for Norwegian and Russian were generated. In the next section, we evaluate them using human judgments.

\section{Evaluation and results}
\label{sec:evaluation}

\subsection{Evaluation setup}

Our ultimate aim is to assign each WUG cluster a human-readable label which is distinct enough to be useful for a linguist or lexicographer. In other words, the label should be helpful in distinguishing one cluster from another\footnote{The clusters themselves can be incorrect, but their fixing is out of scope for this paper.}. To evaluate the labeling methods, we came up with the `\textit{guess the cluster by definition}' task for human evaluation, described below.

For every cluster of a target word, the annotators were shown the label generated for this cluster (without knowing what system it came from) and \textit{two} clusters, represented with five randomly sampled example sentences each (or less, if the cluster has less examples in the WUG). One of those clusters is the one for which the system generated the label and another one is a randomly sampled filler cluster (the filler pairings were generated before the evaluation started and were used throughout the whole process). The humans were asked which cluster is the best fit for the shown label (definition). They also had the options `none of the clusters fits' and `the definition fits both clusters'. Only one choice was allowed. Importantly, the annotators were instructed that the label fits the cluster if it fits the majority of the examples in it. Otherwise, fluency or factual correctness of the definitions were not taken into account: they only had to be distinctive enough to tell one cluster from another. Our final evaluation metrics is accuracy for all the clusters, where the system label is considered to be `correct' if and only if the human chose as the best fit the same cluster that the system generated the label for. Our additional metrics are the percentages of `fits both' and `fits none' judgments (the less the better). 

German, Russian and Norwegian predictions were evaluated by native speakers of the corresponding languages (the German speaker was familiar with the corresponding WUG data beforehand). English predictions were evaluated by the paper authors who are fluent English speakers. English definitions for the RuDSI dataset were evaluated by three independent annotators (using majority voting as the aggregation method). The resulting  Krippendorff’s $\alpha$ inter-rater agreement is 0.314, which is considered a fair agreement \cite{artstein-poesio-2008-survey}.

\subsection{Experimental results}
Table~\ref{tab:results} shows the evaluation results of the systems for the WUGs we experiment with.  Note that the task we are solving (generating labels or definitions for sets of sentences) is novel in itself, so there are no prior results to compare against. DefGen outperforms both Lesk and GlossReader in  accuracy for all the datasets under comparison (English, German and Russian with English definitions). Interestingly, DefGen is slightly more prone to producing too general definitions which fit both clusters, but much more rarely produces irrelevant definitions which fit none of the clusters (for Lesk, more than half of the definitions are like this). This is probably the reason for the higher DefGen accuracy score.

For the rest of the WUGs (Norwegian with Norwegian and English definitions and RuDSI with Russian definitions), we evaluated only DefGen, to make sure its performance does not drop for some reason. The results for Norwegian are on the same level or higher, but the accuracy of Russian definitions for RuDSI are much lower than the English definitions generated with the same method. This partially stems from an increased number of `fits none' judgments (the definition is irrelevant for both clusters). We observed low ROUGE-L performance of the Russian DefGen before (subsection~\ref{subsec:defgen}), so it does not come as a surprise. 

However, this degradation can be caused by some properties of a particular dataset, as is suggested by the difference between English and Norwegian definitions for the Norwegian clusters. As mentioned before, there are two sets of Norwegian DWUGs. For Norwegian-2, the English definitions are more accurate than the Norwegian ones (88.24 vs 76.47), same tendency as for Russian. But for Norwegian-1, it's vice versa (60.87 vs 73.91). Thus, the performance of English definitions for Norwegian examples varies greatly: can be excellent for one set of WUGs but poor for another. At the same time, the performance of Norwegian definitions is fairly good for both. We conjecture that the definitions in WUG-native languages can be more robust and less influenced by peculiarities of specific target words. If this is true, it would mean that RuDSI simply happened to be a `lucky' set of WUGs for English definitions, same as Norwegian-2. To test this hypothesis, another Russian dataset is needed, so this is left for future work.

\begin{table}
\centering

\begin{tabular}{@{}l|l|ll@{}}
\toprule
\textbf{System} & \textbf{Accuracy} & \textbf{Fits both} & \textbf{Fits none} \\
\midrule
\multicolumn{4}{c}{English DWUG, English definitions} \\
\midrule
Lesk &  21.67 & 5.00\% & 53.33\% \\
GlossReader & 50.00 & 9.17\% & 37.50\% \\
DefGen & \textbf{69.17} & 10.83\% & 11.67\% \\
\midrule
\multicolumn{4}{c}{German DWUG, English definitions} \\
\midrule
GlossReader & 53.68 & 13.68\% & 27.37\% \\
DefGen & \textbf{57.89} & 16.84\% & 12.63\% \\
\midrule
\multicolumn{4}{c}{RuDSI, English definitions} \\
\midrule
GlossReader & 64.10 & 10.26\% & 17.95\% \\
DefGen & \textbf{71.79} & 15.38\% & 2.56\% \\
\midrule
\midrule
\multicolumn{4}{c}{Norwegian-1 DWUG, English definitions} \\
\midrule
DefGen & 60.87 & 13.04 & 21.74 \\
\midrule
\multicolumn{4}{c}{Norwegian-2 DWUG, English definitions} \\
\midrule
DefGen & 88.24 & 5.88\% & 5.88\% \\
\midrule
\multicolumn{4}{c}{Norwegian-1 DWUG, Norwegian definitions} \\
\midrule
DefGen & 73.91 & 4.35\% & 21.74\% \\
\midrule
\multicolumn{4}{c}{Norwegian-2 DWUG, Norwegian definitions} \\
\midrule
DefGen & 76.47 & 11.76\% & 11.76\% \\
\midrule
\multicolumn{4}{c}{RuDSI, Russian definitions} \\
\midrule
DefGen & 48.72 & 7.69\% & 15.38\% \\
\bottomrule
\end{tabular}
\caption{Results of human evaluation in the `guess the cluster by definition' task.}
\label{tab:results}
\end{table}

\subsection{Error analysis}

We have manually classified all erroneous definitions from the English, Norwegian 1 and Russian DWUG by mistake type, the results are shown in the tables~\ref{tab:erreng},~\ref{tab:errnob},~\ref{tab:errrus}. 

\begin{table}[ht]
\centering
\begin{adjustbox}{max width=\columnwidth}
\begin{tabular}{l|l l }
\textbf{Error type} & \textbf{Share} & \textbf{Number}  \\
\midrule
too broad definition & 0.41 & 15 \\
wrong sense & 0.32 & 12 \\
similar words but not a definition & 0.16 & 6 \\
redundant whitespace & 0.08 & 3 \\
ambiguous definition, ambiguous word & 0.03 & 1 \\
repetitions & 0.05 & 2 \\
factual mistake & 0.03 & 1 \\
non-existing word & 0.03 & 1 \\
\midrule
Total & 1.00 & 37 \\
\bottomrule
\end{tabular}
\end{adjustbox}
\caption{Erroneous English definitions by error type. Some definitions were annotated with two error types, for instance, some too broad definitions also suffer from a redundant whitespace.}
\label{tab:erreng}
\end{table}

One of the most widespread problems, which has caused most of `fits both' annotations, is too broad definition. For example, the following definition was generated for the word `relationship': `\textit{The way in which two or more things are connected, or in which two or more things are related.}'. This definition did not allow the annotator to distinguish the clusters containing usages related to relationship among entities or among people. The same word causes `fits both' in Norwegian: `forhold' (`relationship') is defined as `\textit{forbindelse, samsvar mellom ulike faktorer}' (`connection, agreement between different factors'). A similar problem was also reported by the annotator of the German data: `\textit{Concreteness vs. abstractness seems very important in descriptions: Sometimes the description could fit to both clusters, depending on whether it is read concretely or abstractly (metaphorically), e.g. `beimischen', `to mix (something) together'. Mixing could be done e.g. by mixing chemicals or ideas or thoughts. Similarly with `abdecken: to cover or supply}'.

\begin{table}[ht]
\centering
\begin{adjustbox}{max width=\columnwidth}
\begin{tabular}{l|l l }
\textbf{Error type} & \textbf{Share} & \textbf{Number}  \\
\midrule
wrong sense & 0.5 & 3 \\
similar words but not a definition & 0.17 & 1\\
too broad definition & 0.17 & 1 \\
repetitions & 0.17 & 1 \\
wrong preprocessing & 0.17 & 1 \\
\midrule
Total & 1.00 & 6 \\
\bottomrule
\end{tabular}
\end{adjustbox}
\caption{Erroneous Norwegian definitions by error type. Some definitions were annotated with two error types.}
\label{tab:errnob}
\end{table}

Another source of too broad definitions is joining multiple meanings of a word into one definition by semicolon (there can be also a redundant space added before this semicolon; we consider such cases to be a punctuation error. A possible reason is that the training data were pre-tokenized and joined by a space). For example, `plane' is defined as `\textit{A flat surface, without slope, tilts, or indentations ; a level surface.}' which merges its everyday and mathematical meanings.

Some definitions use ambiguous words so that the resulting sense description does not help to distinguish between clusters. For example, `\textit{Something used to support or maintain an object, scene, etc.}' for the word `prop'. The word `scene' is also ambiguous and the annotator has selected a wrong cluster because of it. This problem arises even more clearly when the model is trained in other languages with an English prompt. The definitions are often simply direct translations of the target word into English with the same level of ambiguity. For example, the Russian word `\foreignlanguage{russian}{сторона}' (`side') is defined as `one's part, aspect, role'.

Some of `fits both' annotations may be also explained by too granular clustering, but fixing it is beyond of the scope of this paper.

Another frequent problem are definitions that are good themselves, but describe another sense of the word, not the one used in their source clusters. Such definitions result in wrong or `fits none' annotation. They are often generated by usages of the target lemma in phrasal verbs or in metaphorical meaning. For example, a cluster containing usages of `chef d'ouevres' (in the sense of `masterpiece') has generated the definition `\textit{A master of excellence.}' for the target word `chef'.

\begin{table}[ht]
\centering
\begin{adjustbox}{max width=\columnwidth}
\begin{tabular}{l|l l }
\textbf{Error type} & \textbf{Share} & \textbf{Number}  \\
\midrule
wrong sense & 0.5 & 10 \\
repetitions & 0.45 & 9 \\
similar words but not a definition & 0.3 & 6 \\
redundant whitespace & 0.2 & 4 \\
redundant note & 0.2 & 4 \\
too broad definition & 0.15 & 3 \\
redundant parentheses & 0.05 & 1 \\
\midrule
Total & 1.00 & 20 \\
\bottomrule
\end{tabular}
\end{adjustbox}
\caption{Erroneous Russian definitions by error type. Some definitions were annotated with more than one error type.}
\label{tab:errrus}
\end{table}

The third most frequent problem are definitions that are semantically similar to what could be a good definition, but they barely describe the meaning of the cluster. Such cases are called `similar words but not a definition' in our tables. An example is the definition of the word `gas': `\textit{Any fluid substance, especially a natural fluid, which is a mixture of air and nitrogen, usually created by the combustion of natural gasses.}' It sounds much dictionary-like, but is meaningless and also contain a non-existing word `gasses'. The same problem was also reported for German, the annotator has often made his judgments based on `only some aspect that was not present in the other cluster' in such cases. For Norwegian, there is a definition `\textit{liten, sammenhengende fjordformasjon}' (`small, connected fjord formation') generated for the word `rev' (`reef'). Not only is this definition an example of over-fitting to the Norwegian realities, but also, as the annotator has pointed out, `fjordformasjon' means `a formation of the actual fjord, like how the water moves and the sides of the fjord, not the ground in the water in the fjord'.
Sometimes the meaninglessness of the definition is produced by repetitions in it; for example, `pin' is defined as `\textit{To fasten with a pins or pins.}'

Sometimes the definitions with a wrong meaning still carry the sentiment or the style of their source cluster. For example, the definition `\textit{Bravery, courage, chutzpah, or brazenness.}' containing a colloquial word `chutzpah' was generated for the word `ball' in  its colloquial meaning. For German, the annotator could assign the definition to one of the clusters only based on the sentiment while the definition content does not fit otherwise: `\textit{Abgesang: an expression of disapproval, a denial, or rejection. The negative sentiment fits very well with the "downfall" meaning of the word}'. 

Factual mistakes are an important problem. For example, `heel' is defined as `\textit{The part of the human foot above the ankle.}', while it is the part under the ankle. The annotation guidelines\footnote{\url{https://github.com/ltgoslo/wugs_with_definitions/blob/main/human_evaluation/evaluation_guidelines.pdf}} included the instruction to ignore factual mistakes if they allow to distinguish between the clusters, so in fact this type may be not as rare as our current analysis shows. For German, the annotator reported that `\textit{the meaning expressed an antonym or contrast of one of the clusters: e.g. `Schmiere: a highly successful theatrical production' is exactly the opposite: a very bad theatrical play. I judged it still as fitting. Similarly for: Engpass: the property of a  more than adequate quantity or supply; Spielball: invloving active participation; Missklang: any agreeable (pleasing and harmonious) sounds}'.

Some errors are specific for the Russian training dataset and can be fixed by cleaning the dataset. These are the Wiktionary artifacts: redundant footnotes in parentheses (e.g. `\foreignlanguage{russian}{стороны [ 1 ], участники какого-либо конфликтного процессуального действия}') and redundant dictionary notes that seem to occur randomly, e.g. `\foreignlanguage{russian}{только полн. ф.}' (`full form only'), which makes sense for adjectives and participles only, but is generated in the definition of a noun.

To conclude, the definitions produced by the model are not yet ready-to-use. They are in general suitable for semantic change detection or other semantic-related NLP tasks, but not for the full replacement of a human lexicographer.

More definition examples with an attempt to classify them by the error severity into good, borderline and bad ones can be found in~\cref{sec:appendix}.

\section{Conclusion}
We have experimented with several methods for assigning human-readable cluster labels to word usage graphs (both synchronic and diachronic) for multiple languages. Human evaluation showed that conditional generation of such labels from scratch using a fine-tuned \texttt{mT0} language model outperforms two baselines which choose a label from an external closed set of possible definitions (senses). This `DefGen' method generally follows the prior work, but we additionally show that it can generate reasonable labels even for languages on which the model was not specifically fine-tuned. Our definition generation models for English\footnote{\url{https://huggingface.co/ltg/mt0-definition-en-xl}}, Norwegian\footnote{\url{https://huggingface.co/ltg/mt0-definition-no-xl}} and Russian\footnote{\url{https://huggingface.co/ltg/mt0-definition-ru-xl}} are publicly available.

We release enriched versions of the existing WUGs for English, German, Russian and Norwegian. In these versions, clusters of usages are accompanied with labels generated by DefGen, both in English and in the native languages of the datasets. It is our hope that this makes the WUGs easier to use and analyze for linguists and lexicographers: it is now not necessary to browse through all the examples in a cluster to find out what sense of the target word does it correspond to. It is even more important for diachronic WUGs, where senses are changing over time and must be compared historically. Assigning DWUG clusters human-readable definitions is another step towards \textit{explainable} semantic change modeling, thus bridging the gap between NLP and humanities.

We believe our findings support the claim of \citet{giulianelli-etal-2023-interpretable} that generated definitions can be used as convenient and interpretable contextualized representations for a wide range of domains. One can think of using such `definitions as representations' in WSD, WSI or sentiment analysis, to name only a few NLP tasks.

As a future work, we plan to experiment with a DefGen model fine-tuned on definition datasets in several languages at once and with further fine-tuning such a model on a very small amount of manually created examples in other languages. Our preliminary experiments with German are promising (the model is able to generate definitions in German, despite the lack of proper fine-tuning dataset), but more work is needed. It is also planned to generate cluster labels for other existing WUGs  not yet covered in this paper.

\section{Acknowledgments}
Andrey Kutuzov and Nikolay Arefyev have received funding from the European Union’s Horizon Europe research and innovation program under Grant agreement No 101070350 (HPLT). Dominik Schlechtweg has been funded by the research program `Change is Key!' supported by Riksbankens Jubileumsfond (under reference number M21-0021).
We acknowledge the help of Petter Mæhlum in evaluating Norwegian definitions and Pavel Suvorkov in evaluating Russian definitions. 

\section{Bibliographical References}\label{sec:reference}

\bibliographystyle{lrec-coling2024-natbib}
\bibliography{anthology,custom}

\section{Language Resource References}
\label{lr:ref}
\bibliographystylelanguageresource{lrec-coling2024-natbib}
\bibliographylanguageresource{languageresource}

\appendix

\section{Appendix}
\label{sec:appendix}

\begin{figure*}[ht]
\begin{adjustbox}{max width=\textwidth}
\includegraphics{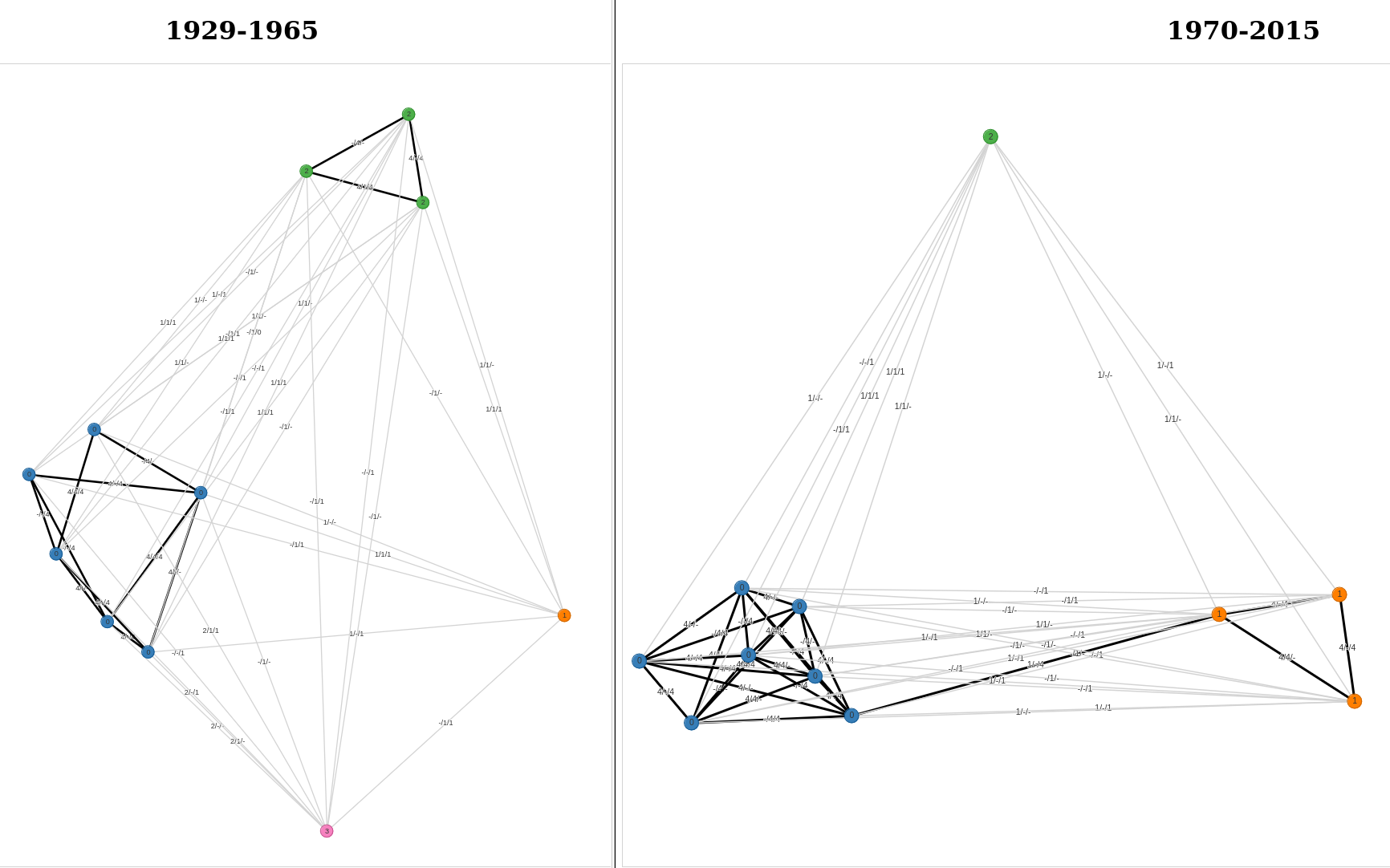}
\end{adjustbox}
\caption{Diachronic word usage graph of the Norwegian word `rev' (`fox'). The blue cluster contains the usages with the sense `fox', the green cluster the usages with the sense `strip of sail', the orange one the usages with the sense `reef' and the rose cluster one usage with the sense `card play'. It can be seen that the sense `card play' no longer exists in the second time period, and the sense `strip of sail' has become rare. }
\label{fig:rev}
\end{figure*}

\begin{table*}[ht]
\centering
\begin{adjustbox}{max width=\textwidth}
\begin{tabular}{l|l l  p{4cm}  p{4cm}  p{4cm} }
\textbf{word} & \textbf{DWUG} & \textbf{sense} & \textbf{prompt} & \textbf{definition} & \textbf{translation} \\
\midrule
Entscheidung & German & decision &  What is the definition of Entscheidung? & The act of making up your mind about something; a decision. & \\
\foreignlanguage{russian}{мир} & Russian & peace & \foreignlanguage{russian}{Что такое мир?} & \foreignlanguage{russian}{полит. соглашение между сторонами о взаимных отношениях} & in politics: agreement between two sides about their mutual relationships \\
\bottomrule
\end{tabular}
\end{adjustbox}
\caption{Good definition examples}
\end{table*}

\begin{table*}[ht]
\centering
\begin{adjustbox}{max width=\textwidth}
\begin{tabular}{l|l  p{2cm}  p{4cm} p{4cm} p{4cm} p{4cm} }
\textbf{word} & \textbf{DWUG} & \textbf{sense} & \textbf{prompt} & \textbf{definition} & \textbf{translation} & \textbf{error} \\
\midrule
rev & Norwegian-1 & fox & Hva betyr rev? & rovdyr som blir brukt som inntektskilde & predator that is used as a source of income & too broad definition\\
rev & Norwegian-1 & reef & What is the definition of rev? & A reefs, especially coral reef walls. &  & grammar mistake\\
\foreignlanguage{russian}{мир} & Russian & community of people & \foreignlanguage{russian}{Что такое мир?} & \foreignlanguage{russian}{перен. совокупность людей, объединённых общностью интересов, интересов, а также их общая совокупность} & metaphorical: a community of
people with shared interests, interests, and also their entirety & repetitions \\
plane & English & & What is the definition of plane? & An aircraft. & & too broad definition \\
\bottomrule
\end{tabular}
\end{adjustbox}
\caption{Borderline definition examples}
\end{table*}

\begin{table*}[ht]
\centering
\begin{adjustbox}{max width=\textwidth}
\begin{tabular}{l|l p{2cm}  p{4cm}  p{4cm}  p{4cm}  p{4cm} }
\textbf{word} & \textbf{DWUG} & \textbf{sense} & \textbf{prompt} & \textbf{definition} & \textbf{translation} & \textbf{error} \\
\midrule
rev & Norwegian-1 & fox & What is the definition of rev? & A rabbit. & & factual mistake\\
rev & Norwegian-1 & strip of sail & Hva betyr rev? & særlig skrevet særlig skrevet ... & -, meaningless repetition of random words & wrong preprocessing or bad generation parameters\\
rev & Norwegian-1 & strip of sail & What is the definition of rev? & A twisting blow ; a punch. & & similar words but not a definition\\
\bottomrule
\end{tabular}
\end{adjustbox}
\caption{Bad definition examples}
\end{table*}

\end{document}